\pdfoutput=1

\documentclass[11pt]{article}

\usepackage[]{emnlp2023}
\usepackage{graphicx}

\usepackage{times}
\usepackage{latexsym}

\usepackage[T1]{fontenc}

\usepackage[utf8]{inputenc}

\usepackage{microtype}

\usepackage{inconsolata}

\usepackage{booktabs} 
\usepackage{listings} 
\usepackage{subcaption} 

\usepackage{colortbl}

\title{Can training neural language models on a curriculum with developmentally plausible data improve alignment with human reading behavior?}

\author{Aryaman Chobey \\
  Colgate University \\
achobey@colgate.edu \\\And
  Oliver Smith \\
  Colgate University \\
  osmith@colgate.edu \\\And
  Anzi Wang \\
  Colgate University \\
  awang1@colgate.edu \\\And
  Grusha Prasad \\
  Colgate University \\
 gprasad@colgate.edu \\}

\begin{document}
\maketitle
\begin{abstract}
The use of neural language models to model human behavior has met with mixed success. While some work has found that the surprisal estimates from these models can be used to predict a wide range of human neural and behavioral responses, other work studying more complex syntactic phenomena has found that these surprisal estimates generate incorrect behavioral predictions. This paper explores the extent to which the misalignment between empirical and model-predicted behavior can be minimized by training models on more developmentally plausible data, such as in the BabyLM Challenge. We trained teacher language models on the BabyLM ``strict-small'' dataset and used sentence level surprisal estimates from these teacher models to create a curriculum. We found tentative evidence that our curriculum made it easier for models to acquire linguistic knowledge from the training data: on the subset of tasks in the BabyLM challenge suite evaluating models' grammatical knowledge of English, models first trained on the BabyLM data curriculum and then on a few randomly ordered training epochs performed slightly better than models trained on randomly ordered epochs alone. This improved linguistic knowledge acquisition did not result in better alignment with human reading behavior, however: models trained on the BabyLM dataset (with or without a curriculum) generated predictions that were as misaligned with human behavior as models trained on larger less curated datasets. This suggests that training on developmentally plausible datasets alone is likely insufficient to generate language models capable of accurately predicting human language processing.

\end{abstract}
\section{Introduction}
The rapidly increasing success of neural language models has resulted in a corresponding increase the use of these models to model human neural and behavioral responses. This research direction has yielded mixed success --- while the surprisal estimates from these language models (i.e., the negative log probability of words given their preceding context) can certainly predict a wide range of neural and behavioral responses \cite{schrimpf2021neural}, there are cases where surprisal estimates from these models generate quantitiatively \cite{huang2023surprisal, van2021single, wilcox-etal-2021-targeted} and even qualitatively \cite{arehalli2020neural, davis-2022-incremental} incorrect predictions. 

To what extent are these incorrect predictions a consequence of the fact that these models are trained on orders of magnitude more data than an average human is exposed to in their lifetime \cite{linzen2020generalization}? Can training these models on more developmentally plausible datasets, such as in the BabyLM challenge \cite{warstadt-et-al-2023-babylm}, bridge the gap between empirical and predicted behavior? Does increased alignment with human behavior come at the cost of success on other NLP tasks? We explore these questions in this paper by training models on the the ``strict-small'' dataset of the BabyLM Challenge ($\sim$10M tokens) and evaluating the models on two types of tasks: first, tasks from the BabyLM challenge designed to test these models' linguistic abilities; second, a large scale reading time dataset of syntactically complex sentences designed to evaluate models' ability to capture aspects of human language processing (SAP benchmark; \citealp{huang2023surprisal}). 

\begin{figure*}
    \centering
    \includegraphics[width=0.8\textwidth]{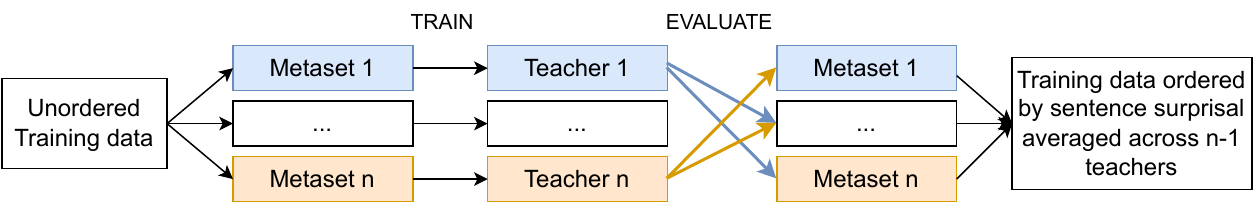}
    \caption{Schematic for how Cross-Review can be used to generate an easy-to-difficult order for the training dataset.}
    \label{fig:cross-review-schematic}
\end{figure*}

Concretely, we explored whether training models on an easy-to-difficult curriculum \cite{elman1993learning} could result in improved performance on the BabyLM suite of challenge tasks and/or an improved fit to human reading behavior in the SAP Benchmark. To design the curriculum, we used the Cross-Review method \cite{xu2020curriculum}: we trained teacher language models on different subsets of the training dataset and then generated sentence level surprisal estimates for held out sentences from each of the teacher models. For every sentence, the surprisal estimates from multiple teachers were averaged together to compute a ``difficulty'' score, which was then used to generate an ordered sequence of training sentences (or ``curriculum''). 

To foreshadow our results, we found that models trained on our curriculum alone performed \textit{worse} on the BabyLM suite of challenge tasks compared to models trained for the same number of steps without a curriculum. However, for a subset of the challenge tasks evaluating models' grammatical knowledge, models which were trained on the curriculum followed by a few randomly ordered training epochs performed better than models trained on the randomly ordered epochs alone. This suggests that while training on the curriculum alone was not sufficient to acquire relevant linguistic knowledge, it might have induced useful biases in the models which made it easier for the models to acquire linguistic knowledge from the training data. 

However, any useful biases that training on the the curriculum might have induced did not result in improved alignment with human reading behavior: models trained on the BabyLM data (with or without the curriculum) had nearly identical performance on the SAP benchmark to each other and to models trained on larger and less curated datasets. This result, along with prior work on training models on child directed speech \cite{yedetore-etal-2023-poor}, suggests that merely training on developmentally plausible data is likely insufficient for bridging the gap between human behavior and language-model predicted behavior.



\section{Background}
Curriculum learning \cite{bengio2009curriculum} refers to training models through a difficulty-based ordering of training examples (i.e. a curriculum), most often ``starting small'' \cite{elman1993learning} from easy examples before progressing to increasingly difficult sentences. In NLP, curriculum learning has been widely used for Machine Translation (e.g., \citealp{platanios2019competence}), but has also been applied more recently to other natural language understanding tasks \cite{xu2020curriculum}. For a survey see \citet{soviany2022curriculum, wang2021survey}.

There are two steps involved in designing a curriculum: assigning a difficulty score to every training example (``difficulty measurer'') and using these difficulty scores to determine the order in which training examples are presented to the model (``training scheduler'') \cite{wang2021survey}. 

\subsection{Difficulty measurer}\label{sec:difficulty-measurer}

Prior work exploring the efficacy of curriculum learning for NLU tasks has used a wide range of properties to compute sentence difficulty such as sentence length, word frequency (or rarity), tree depth, diversity and understandability (for a review, see \citealp{soviany2022curriculum}). None of these properties by themselves can comprehensively capture what makes one sentence more difficult to process or acquire than another. For example, while long sentences are in general more difficult than short sentences, a shorter ambiguous sentence (``the horse raced past the barn fell'') is more difficult to process than a longer unambiguous one (``the horse which was raced past the barn is the same horse that fell''). Given the complex ways in which all of the individual properties can interact, a holistic way of combining these properties is likely necessary to generate good measures of sentence difficulty. 

A natural way of combining these properties to compute a difficulty measure is to use a ``teacher'' language model to compute the predictability of words in a sentence: given some context, a good language model will assign lower probabilities to words that result in long continuations with infrequent words and structures and/or continuations that describe improbable or hard-to-understand events. Concretely, in this work we define difficulty of a sentence as the mean surprisal of words in the sentence, as given in equation~\ref{eq:mean_surp}, where $\mathcal{D}$ is difficulty, $L$ is the model being used to compute difficulty, $s_k$ is the $k$-th sentence, and $n$ is the number of words in $s_k$. 

\begin{equation}\label{eq:mean_surp}
    \mathcal{D}(s_k , L) = -\frac{1}{n}\sum_{i=0}^n \log P(w_i \mid w_0 ... w_{i-1}, L)
\end{equation}

There are two issues with estimating sentence difficulty in this manner. First, the difficulty estimates can be inaccurate if the teacher language model is trained on the same data for which difficulty scores are being computed. Second, the difficulty estimates can be affected by noisy idiosyncrasies if they are computed from just one teacher language model. To avoid these two issues, we use the Cross-Review method proposed by \citet{xu2020curriculum}. In this method each teacher is trained on a subset of the data, and then evaluated on all subsets other than the one it was trained on. Therefore if there are n teachers, there are n-1 difficulty scores for each sentence which can then be averaged together for a final difficulty score for the sentence (see Equation~\ref{eq:final-difficulty} and Figure~\ref{fig:cross-review-schematic}). 

\begin{equation}\label{eq:final-difficulty}
    \mathbf{D}(s_k) = \frac{1}{M}\sum_{m=1}^M \mathcal{D}(s_k, m)
\end{equation}

\subsection{Training scheduler}\label{sec:training-scheduler}
Given a training dataset $E$ in which examples are ordered by difficulty and a training time step $t$, the training scheduler determines the subset of $E$ that the model can be exposed to at $t$. At a broad level there are two types of schedulers: discrete and continuous (see \citealp{wang2021survey} for a more detailed taxonomy of training schedulers). In discrete schedulers, the training proceeds in stages with $m$ training time steps; at all training time steps in a stage $t_i ... t_{i+m}$, the model is exposed to the same subset of $E$.\footnote{This is equivalent to saying that the at any given training stage, the model is trained on $m$ epochs of a subset of $E$.} In continuous schedulers on the other hand, the subset of $E$ that the model is exposed to changes at every training time step.

In this work we use a continuous scheduler proposed by \citet{platanios2019competence}, in which the proportion of $E$ that the model can be exposed to at $t$, $c_{root-p}(t)$, is given by the formula below, where $T$ is the maximum number of training time steps and $c_0$ is the proportion of sentences that the model is exposed in the first time step:

\begin{equation}\label{eq:root-p}
    c_{root-p}(t) = \min\Big(1, \sqrt[p]{t \frac{1-c_0^p}{T} + c_0^p}\Big)
\end{equation}

Our primary reason for using the scheduler above is that it has only three hyperparameters: $c_0$, $T$ and $p$. The authors demonstrate that hyperparameters like warmup steps, which are normally very highly tuned, do not have to be tuned with their scheduler. Given our compute limitations, hyperparameter tuning was infeasible, thus making this approach appealing. 

\begin{table*}[h]
    \centering
     \resizebox{0.75\textwidth}{!}{ 
    \begin{tabular}{llllll}
    \toprule
    Dataset  & Speech? & \# tokens & Proportion & \# sentences & Proportion \\
    \midrule
    CHILDES & Yes & 0.44 M &  4\% & 80K & 9\% \\
    BNC Spoken  & Yes &0.84 M & 9\% & 73.41 K & 8\% \\
    Children's book test & No  & 0.57 M & 6\% & 26 K & 3\% \\
    Children stories & No & 0.34 M & 3\% & 5.72 K & 1\% \\
    Project Gutenberg & No & 0.99 M & 10\% & 91.81 K & 10\%\\
    Open Subtitles & Yes & 3.03 M & 31\% & 470.89 K & 51\% \\
    QED & Yes & 1.03M & 10\% & 91.91 K & 10\% \\
    Simple Wikipedia & No & 1.51 M & 15\% & 48.80 K & 5\% \\
    Switchboard & Yes & 0.11 M & 1\% & 11.09 K & 1\% \\
    Wikipedia & No & 0.99 M & 10\% & 19.35 K & 2\% \\
    \midrule
    Total & & 9.87 M &  & 918.98 K &  \\
    \bottomrule
    \end{tabular}
    }
    \caption{Number of tokens and sentences in each of the sub-datasets in the BabyLM ``strict-small'' datasets. The number of tokens are based on a BPE tokenizer we trained and we exlcude tokens from lines with just a tab or space. Therefore the numbers are slightly different from those in the BabyLM call for papers.}
    \label{tab:dataset-overview}
\end{table*}
\begin{figure*}
    \centering
    \includegraphics[width=0.75\textwidth]{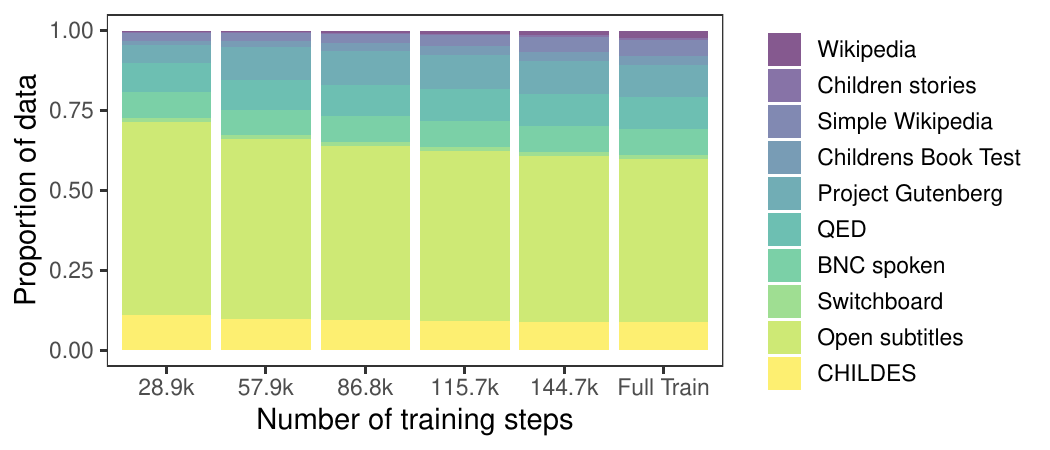}
    \caption{\label{fig:root10-by-domain}Proportion of sentences from each of the sub-datasets in our training curriculum for every 28937 training steps (i.e. equivalent to 1 epoch in the random models), as well as the proportions in the entire training dataset.}
    
\end{figure*}
\section{Designing the curriculum}

\subsection{Datasets}

We trained our random baselines and designed our curriculum using the datasets provided in the ``strict-small'' track of the BabyLM challenge. The data for this track was made up of 10 datasets, with a total of about $\sim$10M tokens and $\sim$920K sentences (where sentences were defined as sequences separated by a new line character). As specified in the BabyLM call for papers, the relative distribution of the ten datasets at the token level was intended to be developmentally plausible -- for example, about 55\% of all the tokens in the ``strict-small'' dataset comes from transcribed speech, and another 19\% of the tokens come from stories (see Table~\ref{tab:dataset-overview}). 

While the BabyLM challenge datasets were constructed at the \textit{token level}, we designed our curriculum at the \textit{sentence level}, where we defined sentences as sequences separated by a new line character. We did this because it was  more straightforward to sort the training dataset based on the difficulty of entire sentences; creating a token-level curriculum would require specifying an additional mechanism for ensuring that contextual integrity was maintained. The relative distribution of the ten datasets at the sentence level was very different from the relative distribution of tokens (see Table~\ref{tab:dataset-overview}). Specifically, the proportion of more ``complex'' datasets (such as Wikipedia and Simple Wikipedia) was much lower at the sentence level than at the token level. We discuss the consequence of these distributional differences in \S~\ref{sec:discussion}.

\subsection{Computing sentence difficulty}
As discussed in \S~\ref{sec:difficulty-measurer}, we used the Cross-Review method proposed by \citet{xu2020curriculum} to compute the difficulty of every sentence in the training dataset. We divided the training dataset into five metasets, each with approximately the same number of tokens and number of sentences. Then, we used the neural-complexity codebase \cite{van-schijndel-linzen-2018-neural}\footnote{https://github.com/vansky/neural-complexity} to train five LSTM teachers on each of these metasets. 

Our LSTM teachers each had two hidden layers with 200 units in each layer. Training sentences were pre-tokenized using a BPE tokenizer that we trained (described in \S~\ref{sec:tokenizer}) and were passed to the teacher models in 20 batches. They were trained until their validation loss did not improve for three epochs, or until they reached 100 epochs. All teachers converged within 67 epochs, with the fastest teacher converging in 54 epochs. 

We then evaluated each of the teacher LSTMs on all metasets except the one they were trained on, and then used the resulting surprisal values to compute the difficulty of every sentence in the training dataset (see Equation~\ref{eq:final-difficulty} and Figure~\ref{fig:cross-review-schematic}).

\paragraph{Why use LSTM teachers?} We trained LSTM language models instead of transformers because prior work has demonstrated that for datasets with 4 million tokens or less, such as our metasets, LSTM language models outperform their transformer counterparts \cite{hu-etal-2020-systematic}, and therefore would make better ``teachers''. Note, we did not use state-of-the-art language models as our teachers because of the constraints of the \textit{strict-small track} of the BabyLM challenge. 

\subsection{Creating the training dataset}
As discussed in \S~\ref{sec:training-scheduler}, we use the training scheduler proposed by \citet{platanios2019competence} which has three hyperparameters (see Equation~\ref{eq:root-p}): the initial competence ($c_0$), the total number of training steps ($T$) and the root value ($p$). Following \citet{platanios2019competence}, we set the value of $c_0$ to $0.01$. We set the value of $T$ to be $150001$ because our random baseline (described in \S~\ref{sec:modeltypes}) achieved the highest validation perplexity after $144685$ training steps (i.e., after 5 epochs).\footnote{It was $150001$ instead of $150000$ because of an error.} We set the value of $p$ to be $10$ after some experimentation because for values of $p$ lower than that, the complex domains in our training dataset (such as Wikipedia) were very underrepresented (see Figure~\ref{fig:by_domain_allroots} in the Appendix). Then, for every batch, we sampled 32 sentences from the subset of sentences that the model can be exposed to at the current time step as determined by Equation~\ref{eq:root-p}.



\section{Training}\label{sec:training}

\subsection{Model architecture}
For our target models we use the OPT 125M architecture \cite{zhang2022opt}. This decoder-only transformer architecture consists of 12 layers with 12 attention heads each, an embedding size of 768 and a context length of 2048 tokens. We additionally use a final 0.2M layer with a  causal language modeling head.

\subsection{Tokenization and batching}\label{sec:tokenizer}
Since the BabyLM challenge does not permit the use of pretrained tokenizer, we trained a BPE tokenizer on the training dataset with a vocabulary size of 50272 (the same as was used in the original OPT models). Like in the GPT-2 \cite{radford2019language} tokenizer implementation, we do not significantly normalize or pre-tokenize the tokenizer training data. For the batching process, the tokenizer truncates sequences longer than 128 tokens, and returns the overflowing tokens as a separate sequence; only about 2\% of our training examples were truncated. We used batch size of 32 with dynamic padding. The entire training dataset was divided into 28937 batches or training steps per epoch. 

\subsection{Model types}\label{sec:modeltypes}
\paragraph{Random baseline:} A randomly initialized OPT 125M model trained on our training dataset without any curriculum for up to 8 epochs. We present results from two baselines: the checkpoint after the 5th epoch (RandOPT 5ep; 144685 training steps) which had the best validation loss, and the last checkpoint (RandOPT 8 ep; 231496 training steps). 

\paragraph{Curriculum only model:} A randomly initialized OPT 125M model trained on our entire curriculum (CurrOPT; 150001 training steps). 

\paragraph{Curriculum + Finetuning:}
The checkpoint of the CurrOPT model after it was trained on 144685 steps (i.e., same number of steps as the RandOPT 5ep model) further ``finetuned'' on the entire randomly ordered training dataset for upto 5 additional epochs. We present results from the checkpoint after 3 finetuning epochs (CurrOPT\_ft 3ep; 231496 training steps, same as RandOPT 8ep) and the checkpoint after 5 finetuning epochs (CurrOPT\_ft 5ep; 289370 training steps).

\subsection{Training procedure}
We use an AdamW optimizer \cite{loshchilov2017decoupled} with $\beta_1$ and $\beta_2$ set to $0.9$ and $0.95$ respectively. We use a weight decay and dropout of $0.1$, and clip gradient norms at $1.0$.  For our random baseline we use a linear learning rate schedule and use a warmup of $\sim$5\% of our maximum training steps. As discussed in \S~\ref{sec:training-scheduler} we do not use warmup for our curriculum models. Due to our considerably smaller pre-training corpus we do not implement  the several mid-flight changes to learning rate and gradient clipping employed by \citeauthor{zhang2022opt} (as an adhoc response to training instability) over the course of their significantly longer training run.

\section{Evaluation}\label{sec:eval}

We evaluate our models on the three challenge sets included in the BabyLM challenge -- BLiMP \cite{warstadt-etal-2020-blimp-benchmark}, (Super)GLUE \cite{wang-etal-2018-glue, wang2019superglue} and MSGS \cite{warstadt-etal-2020-learning} --- as well as on the SAP Benchmark \cite{huang2023surprisal}.

\paragraph{BLiMP and BLiMP supplement}
The Benchmark of Linguistic Minimal Pairs (BLiMP) probes the linguistic knowledge that a language model encodes by measuring how often the model accurately assigns higher probabilities to words in minimally different grammatical and ungrammatical sentences. The original dataset contains minimal pairs for 12 different linguistic phenomena probing English morphology, syntax and semantics. The BabyLM challenge supplements this dataset with five additional linguistic phenomena targeting discourse level acceptability as well as other syntactic phenomena (such as question formation). 

\paragraph{SuperGLUE}
The General language Understanding Evaluation (GLUE) benchmark and its successor SuperGLUE are challenge sets that are designed to evaluate models' general purpose natural language understanding. The BabyLM challenge includes tasks from GLUE (COLA, SST2, MRPC, QQP, MNLI, QNLI, RTE), three tasks from SuperGLUE (BoolQ, RTE and WSC), as well as an additional task (Multimodal NLI). Unlike BLiMP which largely evaluates grammatical knowledge, the SuperGLUE tasks are designed to evaluate higher level linguistic abilities such as sentiment analysis, inference, causal reasoning, coreference resolution, question answering, paraphrasing, etc.


\paragraph{MSGS}
The Mixed Signals Generalization Set is a diagnostic set used to evaluate how models solve an ambiguous classification task that can be solved using either linguistic features or surface features. The MSGS set contains five surface features and four linguistic features, resulting in 20 ambiguous classification tasks. There are also 9 control tasks to evaluate how well models can classify each of the features in an unambiguous context. The BabyLM challenge uses three linguistic features (syntactic position, syntactic construction, and syntactic category) and two surface features (lexical content and relative position), thus resulting in six ambiguous classification tasks.

\begin{table*}[]
    \centering
    \resizebox{0.75\textwidth}{!}{  
    \begin{tabular}{p{1.5cm}p{2.5cm}p{1.5cm}p{1.5cm}p{1.5cm}p{1.75cm}p{1.5cm}p{1.5cm}}
    \toprule
       \textbf{Dataset} & \textbf{Task}  & \textbf{RandOPT 5 ep} & \textbf{RandOPT 8 ep} & \textbf{CurrOPT} & \textbf{CurrOPT\_ft 3 ep} & \textbf{CurrOPT\_ft 5 ep}  \\
       \midrule
       & & 114685 steps & 231496 steps & 150001 steps & 231496 steps & 289370 steps \\
       \midrule
       BLiMP  & \cellcolor{orange!25}Anaphor Agr. & 75.72 & \textbf{86.71} & \cellcolor{red!25}70.35 & \cellcolor{red!25}72.59 & 75.05 \\
       & Agr. structure & 66.09 & 67.85 & \cellcolor{green!40}67.8 & \cellcolor{green!40}\textbf{70.44} & 70.04 \\
       & \cellcolor{teal!25}Binding & 69.19 & 66.83 & \cellcolor{green!40}69.23 & \cellcolor{green!40}69.29 & \textbf{72.72} \\
       & Control/Raising & 63.65 & 65.89 & \cellcolor{red!25}63.57 & \cellcolor{green!40}66.81 & \textbf{68.58} \\
       & D-N Agr. & 72.33 & 74.64 & \cellcolor{red!25}72.05 & \cellcolor{green!40}76.58 & \textbf{78.6} \\
       & Ellipsis & 52.71 & 52.89 & \cellcolor{green!40}53 & \cellcolor{green!40}\textbf{61.66} & 55.43 \\
       & Filler-gap & 72.84 & 73.5 & \cellcolor{red!25}72.74 & \cellcolor{green!40}74.4 & \textbf{75.18} \\
       & \cellcolor{orange!25}Irregular forms & \textbf{82.39} & 71.04 & \cellcolor{red!25}82.34 & \cellcolor{green!40}80.97 & 81.27 \\
       & Island effects & 52.06 & 57.21 & \cellcolor{green!40}57.1 & \cellcolor{green!40}\textbf{64.13} & 62.48 \\
       & \cellcolor{teal!25}NPI licensing & 47.46 & 41.59 & \cellcolor{red!25}38.76 & \cellcolor{green!40}45.52 & \textbf{48.25} \\
       & Quantifiers & 55.69 & 64.4 & \cellcolor{red!25}52.81 & \cellcolor{green!40}67.03 & \textbf{67.34} \\
       & Subj-Verb Agr & 63.92 & 64.43 & \cellcolor{red!25}58.84 & \cellcolor{green!40}64.73 & \textbf{65.55} \\
       \midrule
       BLIMP & \cellcolor{orange!25}Hypernym & \textbf{50.58} & 49.19 & \cellcolor{red!25}48.95 & \cellcolor{red!25}47.91 & 47.21 \\
       Supplement & Congr. (easy) & 48.44 & 51.56 &  \cellcolor{green!40}50 & \cellcolor{green!40}\textbf{53.12} & \textbf{53.12} \\
       & \cellcolor{orange!25}Congr. (tricky) & \textbf{36.97} & \textbf{36.97} & \cellcolor{red!25}36.36 & \cellcolor{red!25}36.36 & 36.36 \\
       & \cellcolor{orange!25}Subj-Aux Inv. & 84.92 & \textbf{86.53} & \cellcolor{red!25}72.55 & \cellcolor{red!25}84.58 & 85.02 \\
       & \cellcolor{orange!25}Turn taking & 55 & \textbf{60.71} & \cellcolor{red!25}51.43 & \cellcolor{red!25}55.71 & 57.5 \\
        \midrule
      SuperGlue & COLA & 3.2 & 9.35 & 3.2 &  \cellcolor{green!40}\textbf{9.77} & 8.91 \\
      & SST2 & 83.07 & 83.86 &  \cellcolor{green!40}83.27 &  \cellcolor{green!40}\textbf{85.43} & 83.86 \\
      & MRPC & 73.64 & 75.59 & \cellcolor{red!25}65.50 & \cellcolor{red!25}72.07 & \textbf{80.14} \\
      & QQP & 76.86 & 77.27 & \cellcolor{red!25}74.78 & \cellcolor{green!40}\textbf{77.33} & 76.83 \\ 
      & \cellcolor{orange!25}MNLI & 65.75 & \textbf{67.07} & \cellcolor{red!25}64.63 & \cellcolor{red!25}65.18 & 65.3 \\
      & \cellcolor{orange!25}MNLI-MM & 65.88 & \textbf{66.31} & \cellcolor{red!25}65.58 & \cellcolor{red!25}66.2 & 66.22 \\
      & \cellcolor{teal!25}QNLI & 60.63 & 59.84 & \cellcolor{red!25}59.54 & \cellcolor{green!40}\textbf{61.33} & 60.98 \\
      & \cellcolor{teal!25}RTE & 51.51 & 45.46 & \cellcolor{red!25}47.48 & \cellcolor{green!40}\textbf{53.54} & 48.49 \\
      & \cellcolor{orange!25}BoolQ & 65.15 & \textbf{67.50} & \cellcolor{green!40}66.53 & \cellcolor{red!25}60.30 & 66.81 \\
      & \cellcolor{teal!25}MultiRC & 55.53 & 48.85 & \cellcolor{green!40}\textbf{56.74} & \cellcolor{red!25}46.55 & 47.54 \\
      & WSC & 56.63 & \textbf{61.45} & \cellcolor{green!40}\textbf{61.45} & \textbf{61.45} & \textbf{61.45} \\ 
      \midrule
      MSGS & MV lexical & -100 & -100 & -100 & -100 & -100 \\
      & MV position & -99.95 & -98.39 & \cellcolor{green!40}-99.75 & \cellcolor{green!40}-\textbf{88.76} & -97.62 \\
      &\cellcolor{orange!25}SC lexical & \textbf{0.18} & -57.66 & \cellcolor{red!25}-58.62 & \cellcolor{red!25}-62.46 & -69.88\\
      & \cellcolor{orange!25}SC position & \textbf{-62.68} & -66.39 & \cellcolor{red!25}-62.82 & \cellcolor{red!25}-76.28 & -62.88 \\
      & \cellcolor{orange!25}CR lexical & \textbf{0} & -4.17 & \cellcolor{red!25}-1.7 & \cellcolor{green!40}-2.4 & -1.2 \\
      & \cellcolor{orange!25}CR position & \textbf{-69.49} & -95.59 & \cellcolor{red!25}-87.47 & -\cellcolor{green!40}70.38 & -98.53 \\
      \bottomrule
    \end{tabular}
    }
    \caption{Results for the tasks included in the BabyLM challenge set. Most of the numbers in the table indicate accuracy except for the following cases: MSGS tasks and COLA numbers are Matthew's Correlation Coefficient; MRPC and QQP numbers are F1 scores. Light green cells indicate cases in which CurrOPT and CurrOPT\_ft 3ep performed better than their random counterpart trained on the same number of steps: RandOPT 5ep and RandOPT 8ep respectively. Similarly, red cells indicate cases in which CurrOPT and CurrOPT\_ft 3ep perform worse. Orange cells indicate tasks in which one of the random models ultimately had the best performance. Teal cells indicate tasks in which training the random model on more epochs led to \textit{worse} performance. Bolded numbers indicate the best performance in a task across all five models. For MSGS we interpret ``best performance'' as having the weakest surface bias (i.e., the least negative numbers).}
    \label{tab:babylm_results}
\end{table*}

\paragraph{SAP Benchmark}

The Syntactic Ambiguity Processing (SAP) benchmark is a large scaled reading time dataset for seven different types of syntactically complex sentences. Unlike the other datasets which measure models' linguistic knowledge and ability, this dataset measures whether the models process information as humans do; specifically, whether models and humans are equally surprised by sentences that are grammatical but have complex and infrequent syntactic structures. The data processing pipeline of the SAP benchmark involves three steps: first, estimating \textit{empirical} effects of interest using Bayesian mixed effects models; second, generating predicted reading times from language model surprisal (i.e. negative log probability) values and fitting mixed effects models to estimate \textit{predicted} effects of interest;\footnote{SAP Benchmark uses Bayesian mixed effects models. We use linear mixed effects models because they are less resource intensive to fit and yield nearly identical model estimates.}and third, comparing empirical and predicted effects of interest. The surprisal estimates and reading times are measured at specific \textit{target} words and the following two spillover words. Further details about the different constructions are included in the Supplementary materials. 



\section{Results}


\begin{figure*}
    \centering
     \includegraphics[width=0.68\textwidth]{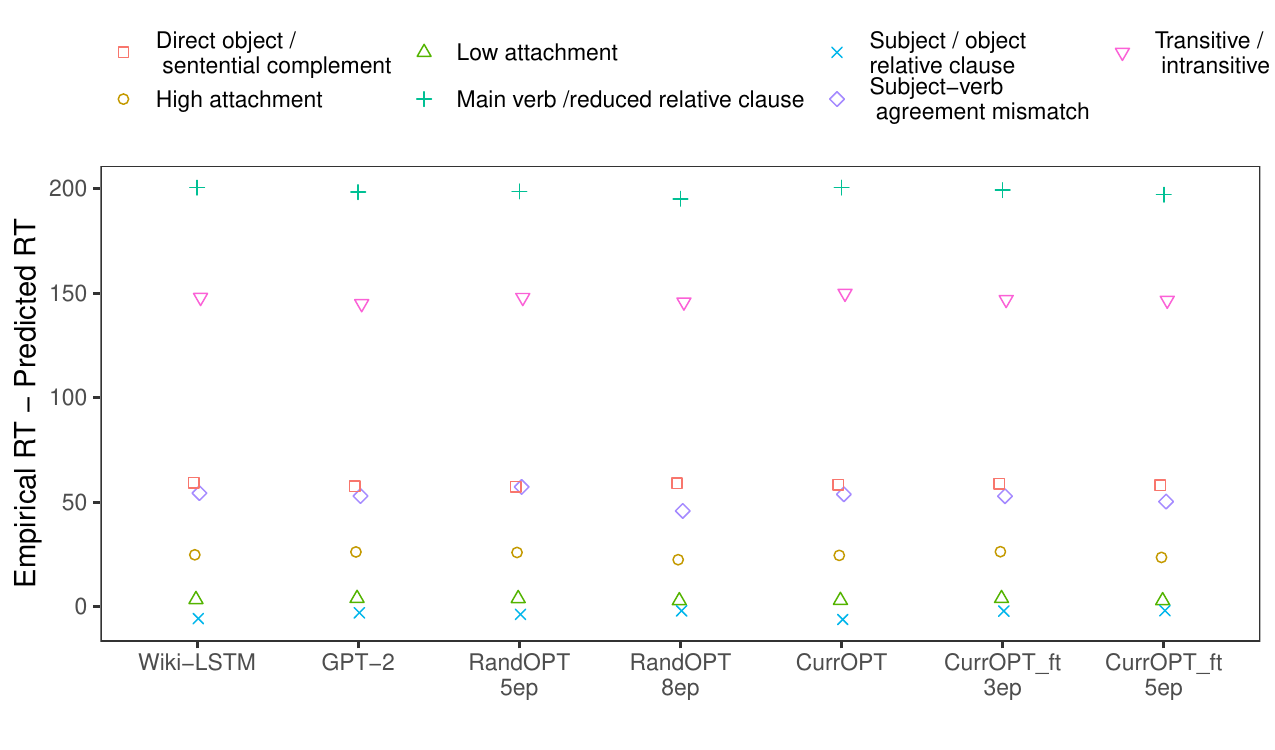}
    \caption{Difference between empirical and model-predicted reading times for the different constructions in the SAP Benchmark. Wiki-LSTM and GPT-2 estimates were from the original paper.}
    \label{fig:sap-results}
\end{figure*}

\subsection{What curriculum was learned?}
The datasets with transcribed speech had the lowest average sentence difficulty scores. Even within transcribed speech, datasets with informal speech (such as child directed speech and subtitles) had lower average difficulty scores than datasets with more formal speech (such as BNC). Additionally, as expected the proportion of transcribed speech steadily decreased over time, as the proportion of written text increased. By the last ``epoch'', the distribution of datasets was very similar to the true distribution (see Figure~\ref{fig:root10-by-domain}), suggesting that the cross-review method we used as our difficulty-measurer was effective, as was the root-10 training scheduler. 

\paragraph{Agreement between LSTM teachers}
For any given sentence, there was a lot of variance in the surprisal estimates across the teachers: the average standard deviation was $113$ bits of surprisal; the mean Spearman rank correlation between any two pairs of teachers was only $0.0009$. This highlights the importance of averaging the surprisal estimates across different teachers to avoid over-fitting to idiosyncrasies of any particular teacher model. 

\paragraph{Other difficulty measures} Figure~\ref{fig:cor-difficulty-measures} plots the correlation between our difficulty measure computed using the cross-review method and two other simpler difficulty measures: average unigram frequency of the words in a sentence and sentence length. Our difficulty measure is moderately correlated with unigram frequency (R = 0.27, p < 0.0001) and highly correlated with sentence length (R = 0.89, p < 0.0001). We also predicted our difficulty measure as a function of unigram frequency and sentence length in a linear regression model and found that unigram frequency explains variance in our difficulty measure over and above sentence length, and together they explain most of the variance in the difficulty measure (adjusted R-squared = 0.93). This suggests that for the specific BabyLM datasets, using cross-review, while effective, might not be necessary: using faster-to-compute measures such as sentence length would have likely resulted in a comparable curriculum. 

\subsection{Training time}
Since our difficulty measure was highly correlated with sentence length, in early stages of training the average sentence length in our curriculum was lower than the average sentence length in early epochs of model training without a curriculum. Since we dynamically padded our sequences, the model trained on our curriculum (CurrOPT) was initially trained on batches consisting of fewer total tokens than the model trained on the unordered data (RandOPT). As a result, in early stages of training, the time taken to train CurrOPT was less than half the amount of time taken to train RandOPT.  As the curriculum progressed, the number of tokens in each batch of CurrOPT approached those in RandOPT causing the training time for CurrOPT to be similar to that of RandOPT. 

\subsection{BabyLM challenge tasks}
On almost all of the tasks, the performance of the model trained on our curriculum (CurrOPT) was worse than the random baseline trained on fewer training examples (RandOPT 5ep). However, when we continued to train CurrOPT on more epochs of the entire training data, the resulting model (CurrOPT\_ft 3ep) performed better than the random baseline trained on the same number of training examples (RandOPT 8ep) on some tasks (see Table~\ref{tab:babylm_results}). Specifically, CurrOPT\_ft 3ep performed better than RandOPT 8ep on most tasks that evaluated models' knowledge of English grammar (e.g., BLiMP, COLA). However, additional training did not seem to help the curriculum models' performance on tasks that required specific lexical knowledge (e.g., irregular forms and hypernyms) or on tasks that required the model to learn more factual information (e.g., MNLI, MNLI-MM and BoolQ). Taken together these results suggest that while training on our curriculum by itself is insufficient to impart the necessary grammatical knowledge, it might induce biases in the model that make it easier for the model to acquire this knowledge from training data. However, there may be limits to the usefulness of these induced biases: training on our curriculum seemed to have some negative impact on the models' ability to acquire nuanced lexical or factual information required to solve more complex tasks like inference or question answering. 

\subsection{SAP Benchmark}
We compared reading times predicted from the surprisal estimates of each of our models, as well as two baselines that were used in the original paper (GPT-2 \cite{radford2019language} and an LSTM model trained on Wikipedia) to the empirical reading times. The difference between predicted and empirical reading times is nearly identical across all models, and very high (greater than 25 ms) for five out of the seven constructions (see Figure~\ref{fig:sap-results}). This difference is not just a result of an incorrect conversion from surprisal to RTs --- we observe qualitatively similar patterns when we look at raw surprisal values (see Figure~\ref{fig:sap_raw_surp} in the Appendix). Thus training on developmentally plausible data (with or without a curriculum) does not result in more human-like processing compared to models trained on less curated written text from the internet. This result aligns with the finding that training on child directed speech does not result in human-like generalization \cite{yedetore-etal-2023-poor}. Taken together these results suggest that merely modifying the training data of language models is unlikely to result in better cognitive models of human language acquisition and processing.  

   

    

\section{Discussion}\label{sec:discussion}
In this paper we explored whether training on a developmentally plausible dataset can improve alignment with human behavior, and whether the improved alignment (if any) comes at the cost of performance on other NLP tasks evaluating different aspects of linguistic competence. We trained models with and without a curriculum on the BabyLM ``strict-small'' dataset and evaluated them on the BabyLM suite of evaluation tasks as well as on a large scale benchmark of reading behavior for syntactically complex sentences (SAP benchmark). 

Drawing on prior work on curriculum learning, we created an easy-to-difficult ordering of the sentences in the training dataset using surprisal values from LSTM teacher language models in a Cross-Review paradigm \cite{xu2020curriculum}, and then used this ordering with a root-10 scheduler \cite{platanios2019competence} to design the training curriculum. This learned curriculum aligned with intuitive expectations for our curriculum --- for example, the proportion of transcribed speech decreased over time, whereas the proportion of written text increased.

An OPT125M causal language model trained on our curriculum (CurrOPT) performed \textit{worse} on most of the tasks in the BabyLM challenge set compared to baselines trained without a curriculum, suggesting that the models were unable to acquire relevant linguistic knowledge from the curriculum alone. Continuing to train CurrOPT on epochs of randomly ordered training data improved performance on most tasks targeting grammatical knowledge, but not on tasks that required more fine-grained knowledge about lexical or factual content. 

\paragraph{Why did training on the curriculum lead to worse performance on some tasks?}
Domains with complex sentences (e.g., Wikipedia) were underrepresented in our curriculum because of our sentence level curriculum: domains like Wikipedia had fewer but longer sentences, and were therefore were less likely to be sampled than sentences from domains with many short sentences (e.g., Open Subtitles). As a consequence there might not have been enough signal in the training data for the models to acquire factual information (which might explain their poor performance on tasks like MNLI and BoolQ) or nuanced lexical representations (which might explain their poor performance on tasks like irregular forms and hypernyms). 

\paragraph{Can training on developmentally plausible data improve alignment with human behavior?}
Crucial to our question, we found that our models which were trained on developmentally plausible data (with or without a curriculum) had nearly identical performance to models trained on less curated larger datasets --- all of the models severely underpredicted the magnitude of processing difficulty in syntactically complex sentences. This suggests that training on developmentally plausible data alone is likely insufficient to bridge the gap between human and model-predicted behavior. 

\paragraph{Limitations and future work}
All of the performance increases that we've discussed were very modest and based on just one model architecture. Therefore further work with additional random runs of the model is required to ensure that the improvements in performance were not just random noise. Similarly repeating the experiments with different architectures for the target and teacher models can shed light on the generalizability of our conclusions. In a similar vein, the conclusions about SAP benchmark results also need to be validated in future work. Specifically, it is necessary to more carefully define what ``developmentally plausible'' means, develop concrete hypotheses about why training on specific datasets might result in better alignment with reading behavior, and test these hypotheses with controlled experiments.

\section{Conclusion}
We designed a surprisal-based curriculum using the developmentally plausible data in the BabyLM strict-small dataset. We found that a model which was first trained on this curriculum and then trained on several additional epochs of the unordered training dataset performed slightly better than a random baseline trained on the same number of examples across a range of NLP tasks. When these models were evaluated on the SAP benchmark, their performance was nearly identical to each other and to that of models trained on larger and less curated datasets. This suggests that merely altering the training data to be more developmentally plausible is unlikely to improve alignment with human behavior.

\section*{Acknowledgements}
This work was supported in part through the Colgate University's ITS HPC resources, services, and staff expertise. We would also like to thank Tom McCoy and Suhas Arehalli and NYU's ITS HPC resources for their assistance with evaluating our models on the BabyLM evaluation tasks, as well as Suhas Arehalli and Forrest Davis for their valuable feedback. 

\bibliography{anthology,custom}
\bibliographystyle{acl_natbib}
\newpage
\appendix

\label{sec:appendix}

\begin{figure*}
    \centering
    \includegraphics[width=0.8\textwidth]{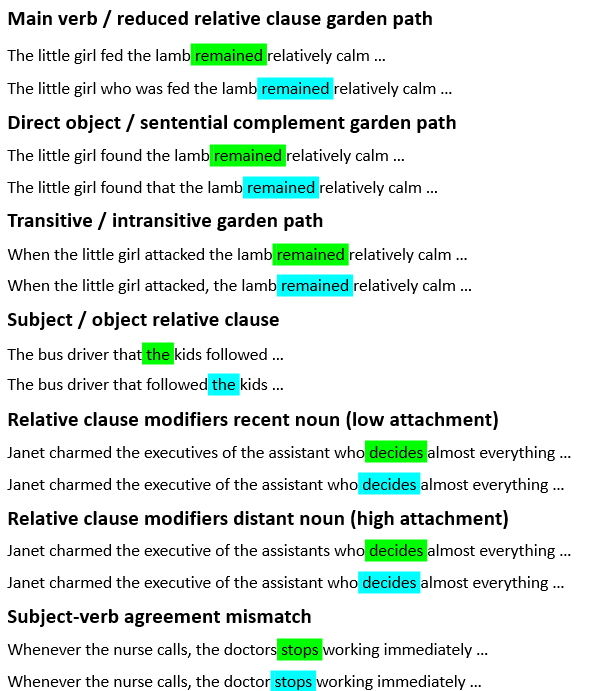}
    \caption{Figure and caption adapted from \cite{huang2023surprisal}. Each sentence pair illustrates a construction tested in SAP Benchmark. An effect of interest is defined as the difference in reading times associated with a disambiguating or ungrammatical word, marked in green, minus the reading time associated with that same word in a context where it is grammatical and does not disambiguate the structure of the sentence, marked in turquoise.}
    \label{fig:enter-label}
\end{figure*}

\begin{figure*}    
\includegraphics[width=0.9\textwidth]{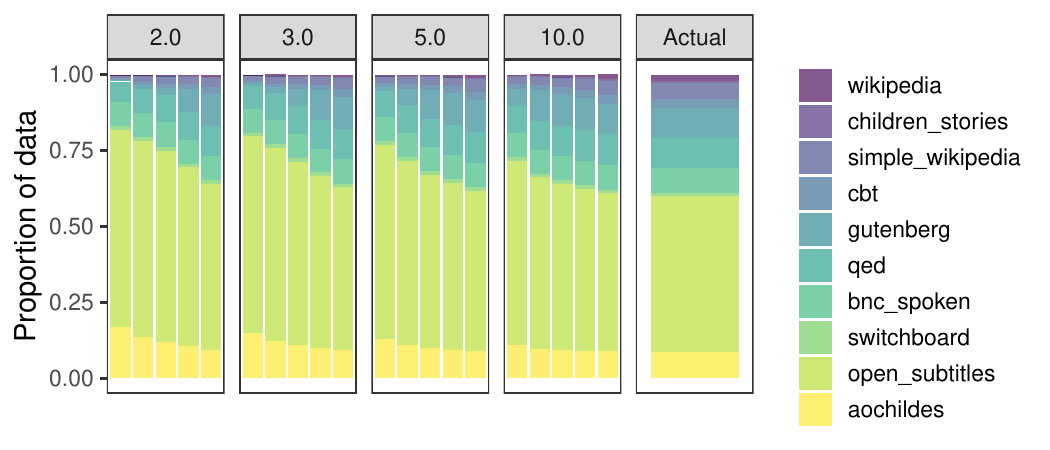}
    \caption{Proportion of sentences from each of the sub-datasets in training curricula with different root values for every 28937 training steps (i.e. equivalent to 1 epoch in the random models), as well as the proportions in the entire training dataset. The curriculum used this is paper is root 10.}
    \label{fig:by_domain_allroots}
\end{figure*}

\begin{figure*}
    \includegraphics[width=0.65\textwidth]{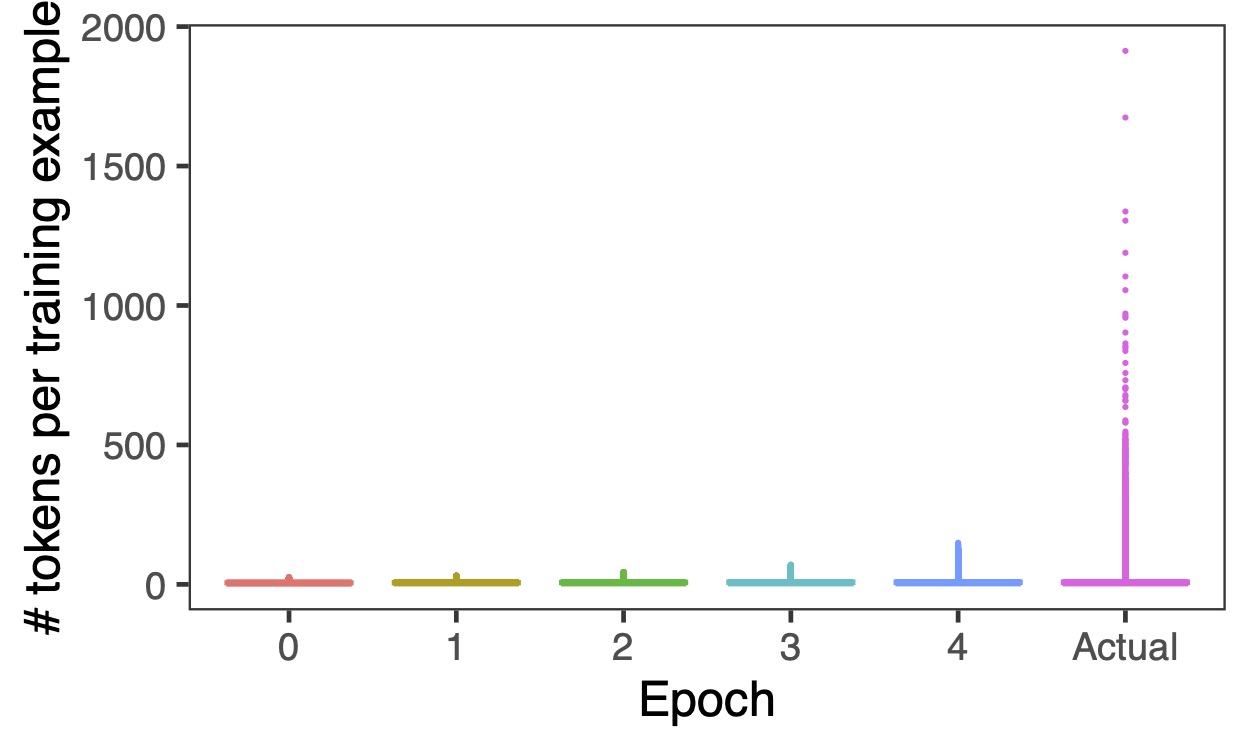}
    \caption{Caption}
    \label{fig:seq_len}
\end{figure*}

\begin{figure*}
    \centering
    \begin{subfigure}[b]{0.4\textwidth}
        \centering
          \includegraphics[width=\textwidth]{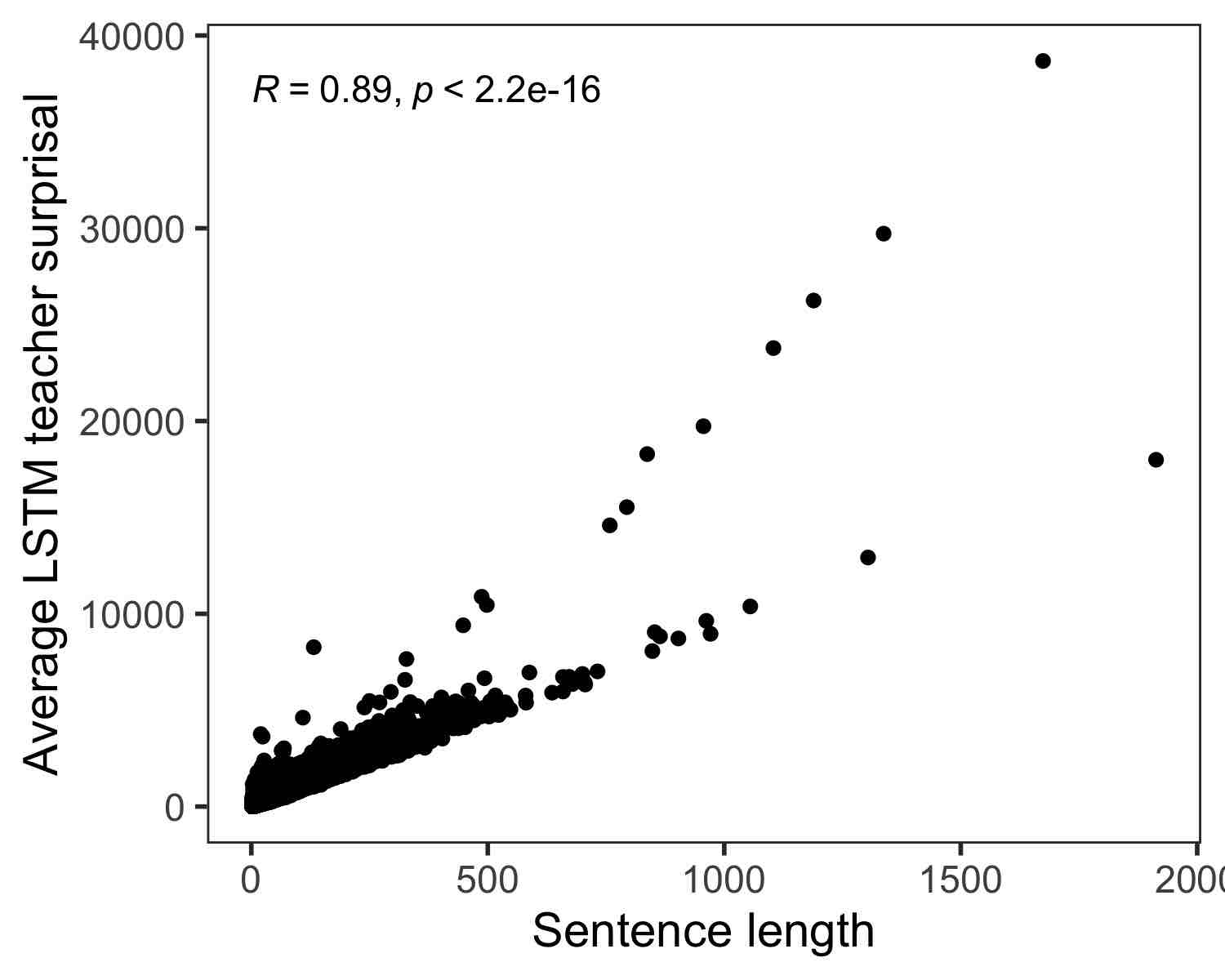}
         \caption{}
         \label{fig:y equals x}
    \end{subfigure}
    \begin{subfigure}[b]{0.4\textwidth}
        \centering
          \includegraphics[width=\textwidth]{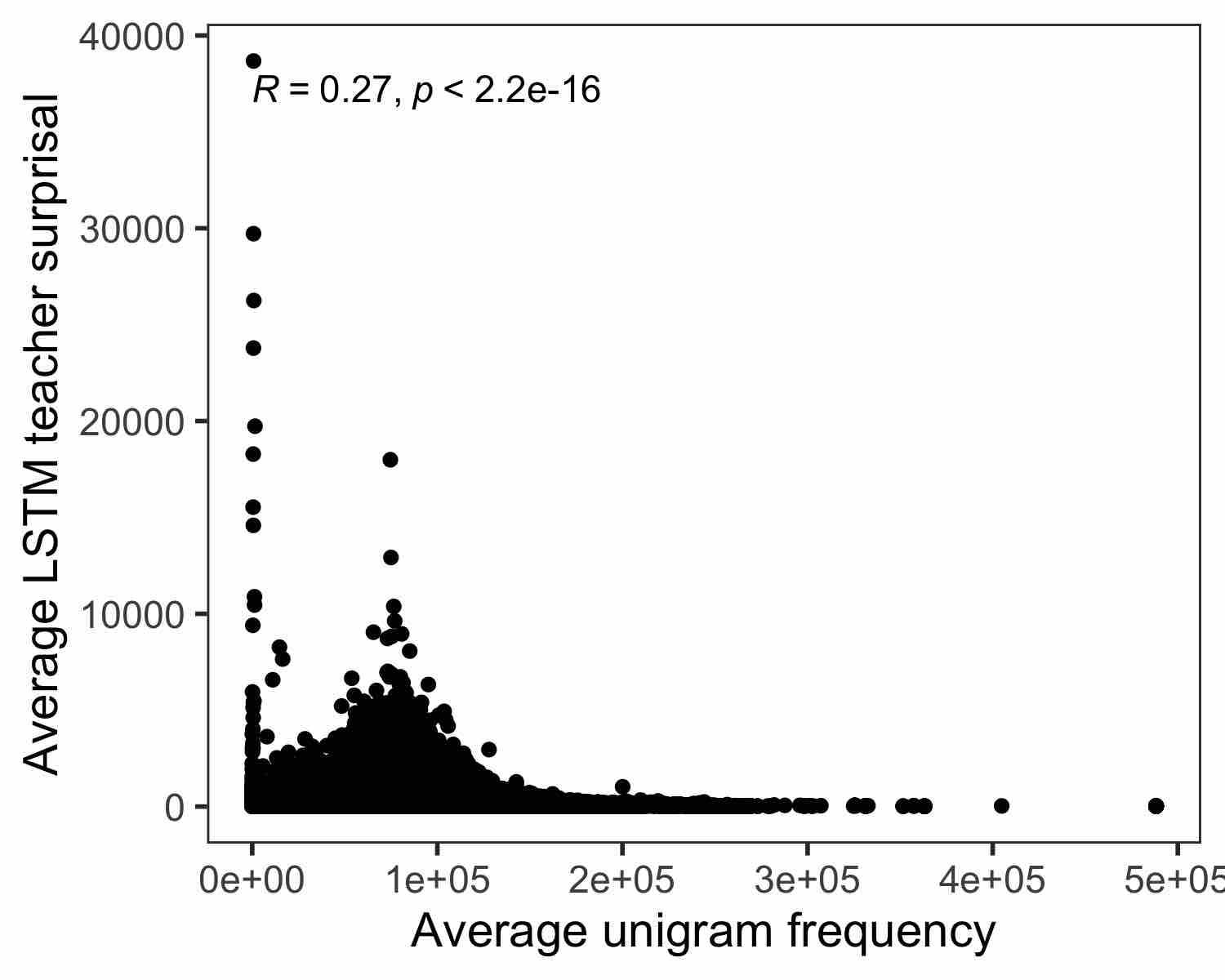}
         \caption{}
         \label{fig:y equals x}
    \end{subfigure}
   
    \caption{Relationship between the order average LSTM teacher surprisal and other difficulty measures. R values indicate the Spearman rank correlation coefficients.}
    \label{fig:cor-difficulty-measures}
\end{figure*}

\begin{figure*}
    \includegraphics[width=0.9\textwidth]{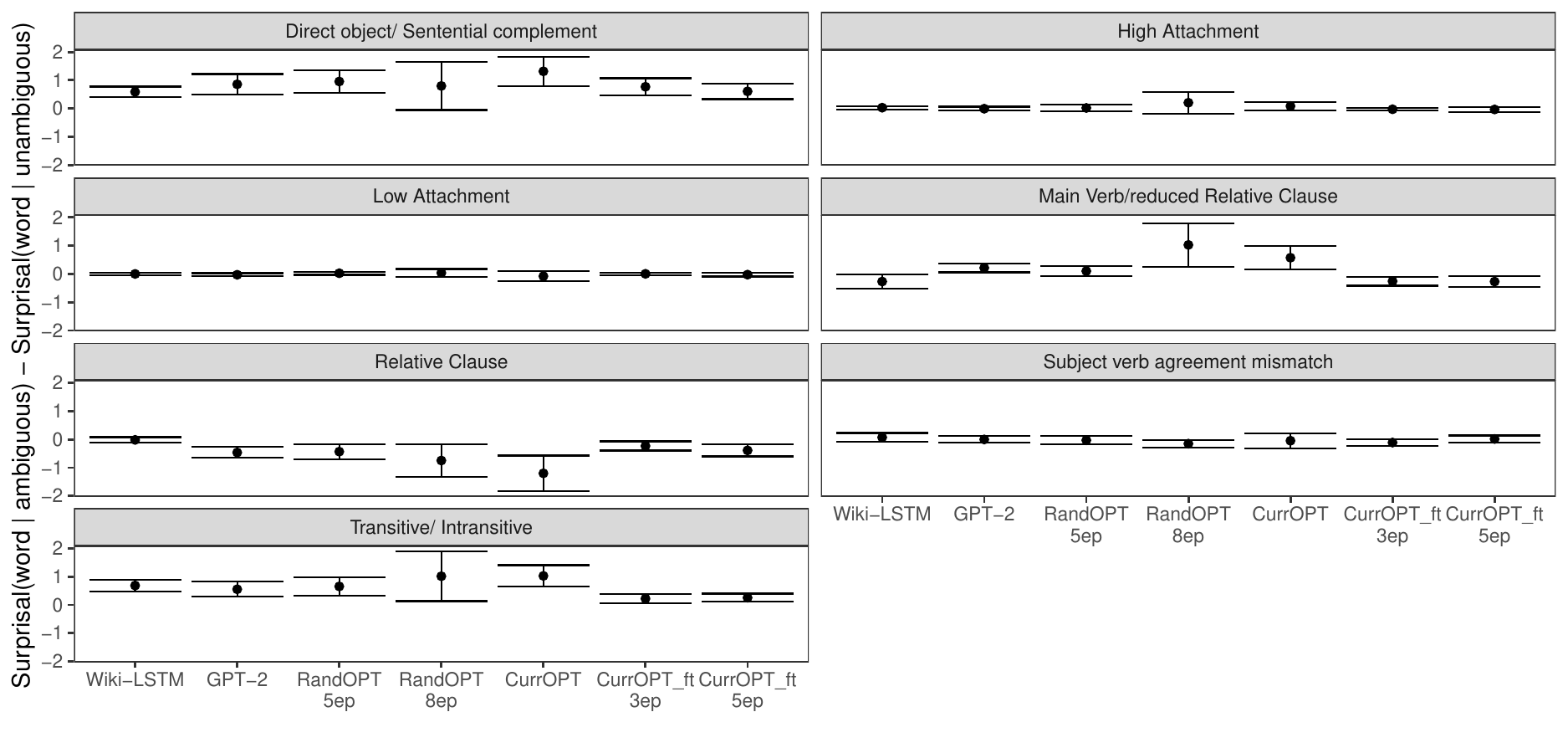}
    \caption{Difference in surprisal value at the target word in ambiguous and unambiguous sentences averaged across all sentences in a construction. Error bars represent 2 standard errors from the mean.}
    \label{fig:sap_raw_surp}
\end{figure*}

\begin{figure*}
    \includegraphics[width=0.98\textwidth]{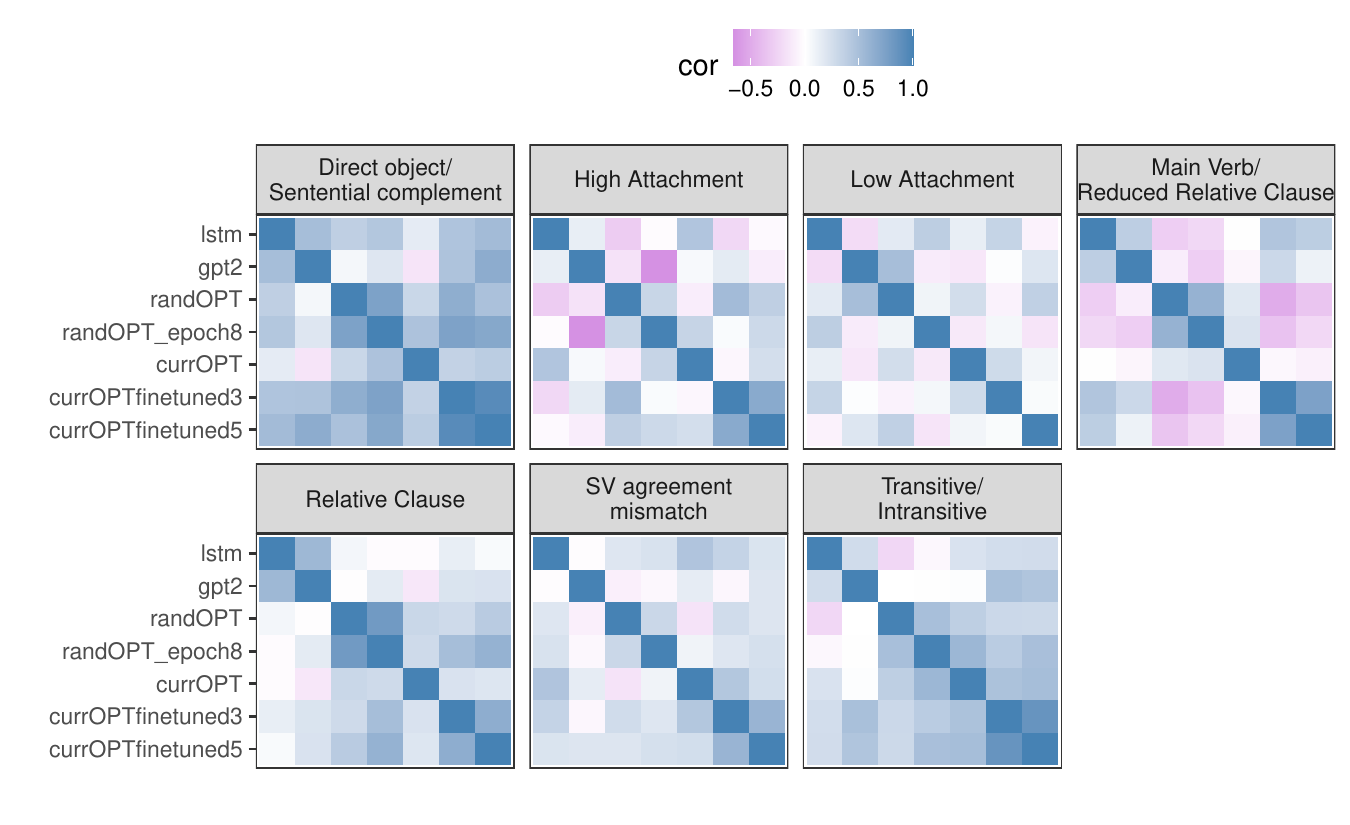}
    \caption{Correlation between model surprisal estimates at the item level. For each model, an item level difference in surprisal at the target word in the ambiguous and unambiguous conditions was computed. This item-level difference was correlated between models using Pearsons's correlation. The relatively weak correlations suggest that even though the models have the same aggregate behavior on the SAP benchmark, their behavior differs at the item level. }
    \label{fig:sap_raw_surp_by_item}
\end{figure*}

\end{document}